\documentclass[a4paper]{article}

\usepackage{cite}
\usepackage[pdftex]{graphicx}
\usepackage[cmex10]{amsmath}
\usepackage{spconf}
\usepackage{flushend}

\newcommand{\figpath}[1]{#1}
\newcommand{\refpath}[1]{#1}

\newcommand{\beq}{\begin{equation}}
\newcommand{\eeq}{\end{equation}}
\newcommand{\bea}{\begin{eqnarray}}
\newcommand{\eea}{\end{eqnarray}}

\newcommand\fig[1]{Figure~\ref{fig:#1}}
\newcommand\tab[1]{Table~\ref{tab:#1}}
\newcommand\eqn[1]{Eq.~(\ref{eq:#1})}

\title{Towards zero-configuration condition monitoring\\ based on dictionary learning}

\name{Sergio Martin-del-Campo$^*$$^\dagger$,  Fredrik Sandin$^*$}

\address{%
    \tabular{c}
                $^*$ EISLAB\\
                 Lule{\aa} University of Technology (LTU)\\
               971~87 Lule{\aa}, Sweden
        \endtabular
        \hskip 0.5in
    \tabular{c}
                $^\dagger$ SKF-LTU University Technology Center\\
                Lule{\aa} University of Technology (LTU)\\
                971~87 Lule{\aa}, Sweden
        \endtabular
}

\begin{document}

\maketitle

\begin{abstract}
Condition-based predictive maintenance can significantly improve overall equipment effectiveness
provided that appropriate monitoring methods are used.
Online condition monitoring systems are customized to each type of machine and need to be reconfigured when conditions change,
which is costly and requires expert knowledge.
Basic feature extraction methods limited to signal distribution functions and spectra are commonly used,
making it difficult to automatically analyze and compare machine conditions.
In this paper, we investigate the possibility to automate the condition monitoring process
by continuously learning a dictionary of optimized shift-invariant feature vectors using a well-known sparse approximation method.
We study how the feature vectors learned from a vibration signal evolve over time
when a fault develops within a ball bearing of a rotating machine.
We quantify the adaptation rate of learned features and find that this quantity changes significantly
in the transitions between normal and faulty states of operation of the ball bearing.
\end{abstract}


\section{Introduction}

Condition monitoring of machine elements is used to detect faults, reduce machine downtime and improve overall equipment effectiveness,
for example by condition-based predictive maintenance.
The requirements on the methods employed to achieve that go beyond fault detection,
in particular in terms of prediction of faults \cite{Yang2002,Yoshioka1984} and detection of abnormal operational conditions.
Early detection and characterization of emerging faults is a challenging problem
because there are many variables that affect the operation of the machine and the characteristics of the fault.
Maintenance operations rely on time and frequency domain features for diagnosis \cite{Yang2002}. 
Expert knowledge is often needed to interpret the features and make decisions, which makes the process difficult to automate.
Furthermore, condition monitoring methods are typically tuned to the application, the operating conditions and the type and location of the fault.
Therefore, such methods are expensive to maintain when machines have varying characteristics and evolve over time,
for example as a consequence of maintenance and repair, which limits the scalability of the approach.
Also, it is difficult to predict all failure modes.
Similarly, approaches based on traditional pattern recognition methods require substantial amounts of labeled training data
and the resulting methods are limited to the conditions for which the method was designed and trained \cite{Randall2011}.

Sparse representation of signals has attracted considerable interest in the last decade \cite{Elad2012,elad2010sparse,Bruckstein2009,Mallat2008}.
One type of sparse representation can be obtained by modeling signals as a linear superposition of noise and 
a small number of atomic waveforms (atoms) of particular shapes, amplitudes and shifts,
so-called shift-invariant sparse coding \cite{Grosse2007,Mailhe2009}.
Using an approach known as dictionary learning the atoms can also be optimized to the signal \cite{elad06ksvd,Mallat2008,Elad2012},
so that each particular atom represents structural features of the signal, which for example are excited by different physical processes.
Such approximations are of increasing interest in signal processing with applications ranging from denoising, source coding, source separation, and signal acquisition.
The problem of finding such sparse representations and optimal atoms is NP-hard in general.
Therefore, suboptimal strategies based on convex relaxation, non-convex (often gradient based) local optimization or greedy search strategies are used in practise.
Liu et al. \cite{Liu2011558} investigate the possibility that faults in a machine can be identified with multiclass linear discriminant analysis using
dictionaries of atoms that are optimized to sets of signals corresponding to different fault conditions of a rotating machine.

In this paper we complement the study by Liu et al. by investigating how one dictionary of atoms changes over time in an online condition monitoring scenario,
where the dictionary is optimized to a continuous vibration signal, measured from a machine, that evolves from a normal state of operation to faulty conditions.
We use a similar implementation of dictionary learning that is suited for online monitoring \cite{MartindelCampo2013},
and vibration signals from the same dataset \cite{Loparo2003}.
%
The work presented here is novel because it focuses on online monitoring and the continuous evolution of an automatically learned dictionary,
rather than supervised learning of multiple dictionaries for each fault condition.
We demonstrate that deviations from the normal state of the machine in principle can be detected via monitoring of the learned dictionary over time.
%
We define an evolution rate for the atoms in a dictionary and demonstrate that this rate decreases to low values after some time of adaptation,
and that it increases significantly when faults are introduced in the system.
The resulting atoms are also useful for further classification and diagnosis of the condition \cite{Liu2011558,MartindelCampo2013}.
We find that some atoms characterize the vibration of the machine in both normal and abnormal operational conditions,
while other waveforms are clearly associated with the faults.
These preliminary results indicate that online monitoring of a learned dictionary is a potentially useful approach to zero-configuration fault detection. 
The approach also provides atoms representing inherent structural features in the signal that can be used for diagnosis and prediction.

\section{Sparse coding and dictionary learning}
\label{sec:Sparse}

The model  \cite{MartindelCampo2013} used here was developed by Smith and Lewicki \cite{smith2006},
and it is inspired by former work on sparse visual coding \cite{Olshausen97}.
Smith and Lewicki discovered that atoms learned from speech data closely resemble cochlear impulse response functions (revcor filters),
which indicates that speech is adapted to the ear \cite{smith2006}.
Our working hypothesis is that features that characterize machines can be learned in a similar manner. 
The model decomposes a signal, $x(t)$, as a linear superposition of noise and atomic waveforms with compact support
\beq
    x(t) = \epsilon(t) + \sum_i a_i \phi_{m(i)}(t - \tau_i).
    \label{eq:superpos}
\eeq
The functions $\phi_m(t)$ are {\em atoms} that represent morphological features of the signal, where $\tau_i$ and $a_i$ indicate the shift (temporal position) and amplitude of the atoms, respectively.
The values of $\tau_i$ and $a_i$ are determined with a matching pursuit algorithm \cite{Mallat1993,Smith2005} and the triple ${m(i), \tau_i, a_i}$ represents one atomic {\it event} (similar to the firing of a receptive-field neuron).
The atoms are optimized in an unsupervised manner by performing gradient ascent on the approximate log data probability \cite{smith2006}
\beq
    \frac{\partial}{\partial \phi_m} \log \left[p(x \mid \Phi)\right] = \frac{1}{\sigma_{\epsilon}^{2}} \sum_i a_{i} (x - \hat{x})_{\tau_{i}},
    \label{eq:probform}
\eeq
where $(x - \hat{x})_{\tau_{i}}$ is the residual of the matching pursuit over the support of atom $\phi_m$ at time $\tau_{i}$ and $a_{i}$ is the atom amplitude.
This is a form of Hebbian learning because adaptation is the result of the continuous activation of the atoms by the input signal.
The stop condition of the matching pursuit algorithm determines the sparseness and signal-to-residual ratio (SRR) of the resulting event-based representation.
Note that the resulting representation is not a linear function of the input signal because the matching pursuit is non-linear.


\begin{figure*}[htb]
\centering
\includegraphics[width=0.75\textwidth]{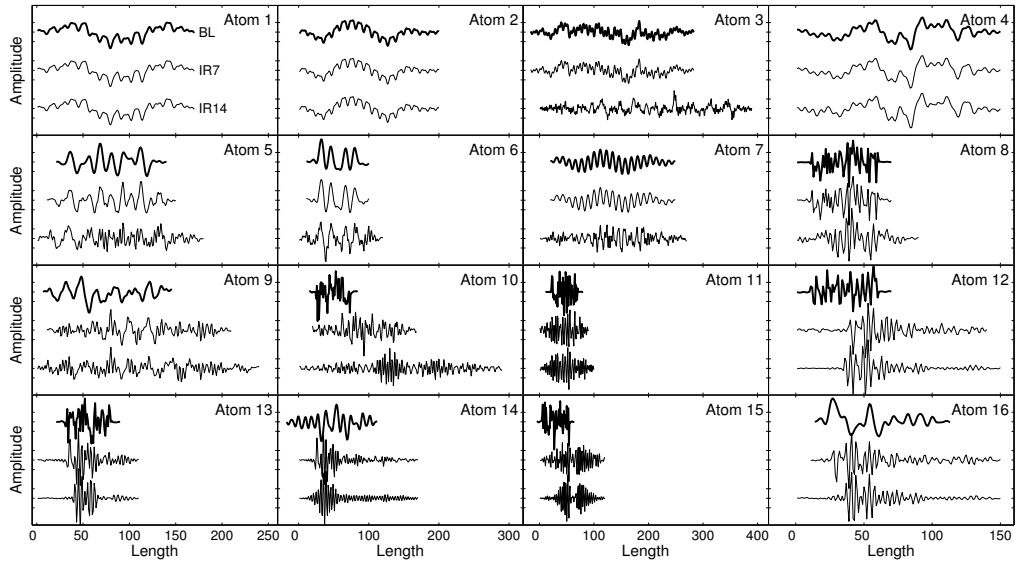}
\vspace*{-0.3cm}
\caption{
Atoms learned from vibration signals corresponding to the BL, IR7 and IR14 cases, respectively.   
The atoms are ordered by ascending center frequencies in the IR14 case.
All atomic waveforms are normalized.
}
\label{fig:DictComp}
\end{figure*}

The set of atoms, $\phi_m(t)$, defines a dictionary, $\Phi$, consisting of $M$ atoms
\begin{equation}
\Phi = \left\{ \phi_1, \dotsc, \phi_{M} \right\}.
\end{equation}
The calculation of $\Phi$ is an iterative process.
The first step is to initialize the dictionary.
In this work we set the initial length of each atom to fifty and sample the initial amplitudes from a Gaussian distribution.
The matching pursuit includes cross-correlation of the signal (residual) with all atoms in the dictionary.
The maximum cross-correlation defines one event, ${m(i), \tau_i, a_i}$, which is subtracted from the signal by subtracting the corresponding waveform, $a_i \phi_{m(i)}(t-\tau_i)$.
The resulting residual is used as input to the next matching-pursuit iteration, and the process continues until the stop condition is reached.
The stop condition can be defined in different ways, for example in terms of the number of events per signal sample (sparsity) or the signal-to-residual ratio.

The problem to learn the dictionary, $\Phi$, is the main challenge and opportunity of this approach,
which makes it fundamentally different from traditional condition-monitoring approaches. 
We seek a dictionary of atoms, $\Phi$, that maximizes the expectation of the log data probability
\beq
\Phi = {\arg\max}_\Phi \langle  \log \left[p(x \mid \Phi)\right] \rangle,
\eeq
where
\beq
p(x \mid \Phi) = \int p(x \mid a,\Phi) p(a) da.
\eeq
The prior of the amplitude, $p(a)$, is defined to promote sparse coding in terms of statistically independent atoms \cite{Olshausen97}.
The integral is approximated with the maximum a posteriori estimate resulting from the matching pursuit.
This results in a learning algorithm that involves gradient ascent on the approximate log data probability defined by \eqn{probform}.
The gradient of each atom in the dictionary is proportional to the sum of residuals corresponding to the matching-pursuit activation of that atom.
The prefactor, $1/\sigma_{e}^{2}$, is the inverse variance of the residual that remains after matching pursuit.
We introduce a {\it learning rate} parameter, $\eta$,
so that \eqn{probform} is modified to
\beq
    \Delta\phi_m  = \frac{\eta}{\sigma_{e}^{2}} {\sum}_{i~:~m=m(i)} a_{i} (x - \hat{x})_{\tau_{i}}.
    \label{eq:probformup}
\eeq
The actual adaptation rates of the atoms also depend on the matching-pursuit activation rate,
which implies that some atoms may adapt slowly or not at all.
Several improvements of this methodology have been proposed, including methods to enforce orthogonality in the matching pursuit.
Such methods improve the reconstruction accuracy significantly for noiseless signals, but the effect on denoising performance is moderate.
Our method is comparable to that used by Liu et al. \cite{Liu2011558}
and is motivated by the relatively low complexity and simplicity of the algorithm,
which allows for online condition monitoring experiments in embedded systems.


We are interested in quantitative changes of the learned atoms resulting from changing conditions in a rotating machine.
Skretting \cite{Skretting2011a} proposes a dictionary distance measure as a means to quantify the similarity between two dictionaries.
This approach is useful for diagnosis purposes but has limitations in an online monitoring scenario because
only a subset of the atoms may change when a fault emerges, possibly resulting in high dictionary similarity.
Therefore, we define the following {\it evolution rate} for each atom
\beq
1-\text{crosscorr}(\phi_a(t), \phi_a(t-\delta)),
  \label{eq:ev_rate} 
\eeq
where $\phi_a(t)$ is an atom of dictionary $\Phi$ at time $t$ and 
$\phi_a(t-\delta)$ is the corresponding atom at a previous point in time,  $t-\delta$.
This quantity is calculated for each atom and it indicates how quickly individual atoms are changing.
A value of zero means no change at all,
while a value close to one means that an atom is uncorrelated with the corresponding atom in the past.

\section{Characterization of rotating machine with fault in rolling element bearing}

We apply the matching pursuit with dictionary learning approach to vibration data from a rotating machine at the bearing data center at Case Western Reserve University \cite{Loparo2003}.
The vibration data was generated with a test rig consisting of an electric motor, a torque transducer, a dynamometer and a ball bearing supporting the motor shaft.
An accelerometer located at the drive end of the motor is used to record the vibration data.
The accelerometer is sampled 12000 times per second.
During data acquisition, the load varies between 0 HP and 3 HP, resulting in a varying motor speed from 1800 to 1730 rpm. 
We consider three different datasets in order to mimic the appearance and growth of a defect in the bearing, 
thereby simulating the evolution of the machine from a normal state of operation to a faulty state of operation.
First, matching pursuit with dictionary learning is applied to 120 minutes of vibration data corresponding to a normal, non-faulty bearing.
This is referred to as the baseline (BL) case and the resulting atoms are illustrated in \fig{DictComp}.
Next, the atoms are further adapted to 120 minutes of data corresponding to a faulty bearing with a 7 mils (0.18 mm) diameter fault on the inner race.
We refer to this as the IR7 case and the resulting atoms are also illustrated in \fig{DictComp}.
Finally, the IR7 atoms are further adapted to 120 minutes of vibration data corresponding to a faulty bearing with a 14 mils (0.356 mm) fault on the inner race (IR14).

The vibration data is processed with our Matlab implementation of Smith and Lewicki's algorithm \cite{smith2006}.
The dictionary initially contains sixteen normalized atoms of length fifty, which are sampled from a Gaussian distribution with zero mean.
Dictionary learning is carried out using a signal window of 5 seconds duration (60000 samples).
The windows are sampled randomly from the different load and rpm cases,
thereby simulating a time-varying load on the rotating machine.
Matching pursuit is stopped at one order of magnitude reduction in the data rate, or at a 12 dB SRR.


The dictionaries resulting from the BL, IR7 and IR14 cases are shown in \fig{DictComp}, each including the sixteen atomic waveforms obtained at the end of a 120 minute adaptation time for each case.
All waveforms are normalized and have the same y-axis scale.
Each panel in \fig{DictComp} illustrates one atom for the BL case (top), IR7 case (middle) and IR14 case (bottom).
Atoms 1, 2 and 4 reach approximately stationary conditions after 120 minutes.
Atoms 9, 10, 12, 13, 14, 15 and 16 change over time and enable distinction of the BL and IR7 cases.
The difference between the IR7 and IR14 cases is evident from the time evolution of atoms 9, 10, 12 and 14. 
Furthermore, the differences between atoms 3, 5, 6, 7 and 8 distinguish the BL and IR14 cases.

\tab{centerfreq} shows the center frequencies of the atoms in the three cases, calculated as the mean value of the power spectral density of each atom. 
\begin{table}[t!]
\medskip
\small
\begin{center}
\renewcommand{\arraystretch}{1.2}
        \begin{tabular}{ccccccc}
        \cline{2-7}
        & \multicolumn{3}{ c }{\textbf{Center freq. [kHz]}} & \multicolumn{3}{ c }{\textbf{Event rate [s\textsuperscript{-1}]}}\\ \cline{1-7}
        \multicolumn{1}{ c }{\textbf{Atom \#}} & \textbf{BL} & \textbf{IR7} & \textbf{IR14} & \textbf{BL} & \textbf{IR7} & \textbf{IR14}  \\ \cline{1-7}
        \multicolumn{1}{ c }{1} & 0.1  & 0.1  & 0.1 & 55 & 0 & 0    \\ \cline{1-7}
        \multicolumn{1}{ c }{2} & 0.1  & 0.1  & 0.1 & 62 & 0 & 0    \\ \cline{1-7}
        \multicolumn{1}{ c }{3} & 1.1  & 1.1  & 0.4 & 99 & 0 & 72    \\ \cline{1-7}
        \multicolumn{1}{ c }{4} & 0.2  & 0.2  & 0.4 & 73 & 0 & 3    \\ \cline{1-7}
        \multicolumn{1}{ c }{5} & 1.0  & 0.9  & 0.5 & 66 & 12 & 49    \\ \cline{1-7}
        \multicolumn{1}{ c }{6} & 0.7  & 0.7  & 0.7 & 64 & 3 & 27    \\ \cline{1-7}
        \multicolumn{1}{ c }{7} & 1.0  & 1.0  & 1.1 & 50 & 4 & 41    \\ \cline{1-7}
        \multicolumn{1}{ c }{8} & 3.9  & 3.9  & 1.9 & 0 & 3 & 63    \\ \cline{1-7}
        \multicolumn{1}{ c }{9} & 0.7  & 1.5  & 1.9 & 58 & 59 & 64    \\ \cline{1-7}
        \multicolumn{1}{ c }{10} & 3.1  & 2.0  & 2.4 & 1 & 116 & 70    \\ \cline{1-7}
        \multicolumn{1}{ c }{11} & 4.4  & 3.2  & 3.1 & 0 & 67 & 76    \\ \cline{1-7}
        \multicolumn{1}{ c }{12} & 2.7  & 3.3  & 3.2 & 0 & 100 & 74    \\ \cline{1-7}
        \multicolumn{1}{ c }{13} & 2.4  & 3.5  & 3.2 & 0 & 99 & 69    \\ \cline{1-7}
        \multicolumn{1}{ c }{14} & 1.1  & 3.6  & 3.4 & 67 & 108 & 69    \\ \cline{1-7}
        \multicolumn{1}{ c }{15} & 2.8  & 3.1  & 3.5 & 0 & 86 & 89    \\ \cline{1-7}
        \multicolumn{1}{ c }{16} & 0.8  & 3.2  & 3.6 & 62 & 107 & 82    \\ \cline{1-7}
        \end{tabular}
\end{center}
\vspace*{-0.3cm}
        \caption{Center frequencies and event rates of learned atoms.}     
        \label{tab:centerfreq}
\end{table}
By calculating the evolution rate (rate of change) of the atoms we notice changes in the characteristics of the rotating machine,
which are associated with the introduction of a fault in the bearing. 
\begin{figure}[t!]
\centering
\includegraphics[width=0.75\linewidth]{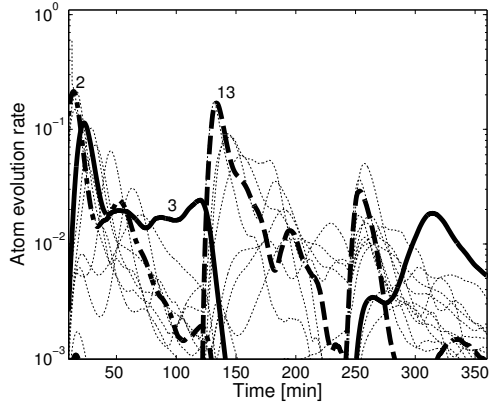} 
\vspace*{-0.3cm}
\caption{
Evolution rate of atoms versus the time in minutes.
The occurrences of the IR7 fault after 120 minutes, and the IR14 fault after 240 minutes affect the evolution rate of some atoms (bold lines) significantly.
}
\label{fig:SimAtom}
\end{figure}
\fig{SimAtom} shows the evolution rate of all the atoms in the dictionary as defined by \eqn{ev_rate} and using $\delta=10$~minutes.
Atom 3 stops evolving when the IR7 case is introduced after 120 minutes, this is represented by the disappearing bold line between 120 and 240 minutes,
which is a consequence of the vanishing event rate, see \tab{centerfreq}.
The center frequency of atom 3 is nearly identical in the BL and IR7 cases, see \tab{centerfreq}.
Atom 3 continues to adapt after 240 minutes when the IR14 case is introduced.
This is in agreement with \fig{DictComp}, which shows that atom 3 is similar for the BL and IR7 cases, while it has a different shape in the IR14 case.
Atom 13 is inactive during the BL case, as indicated by the vanishing event rate in \tab{centerfreq}, but it starts to adapt in the IR7 case and eventually attains an impulse-like shape.
In contrast, atom 2 adapts in the BL case and thereafter remains unchanged, see \fig{DictComp}.
The center frequencies and event rates listed in \tab{centerfreq}, the evolution rate displayed in \fig{SimAtom} and the dictionary illustrated in \fig{DictComp} provide complementary information about the three different operational conditions of the machine.

\begin{figure}[t!]
\centering
\includegraphics[width=0.85\linewidth]{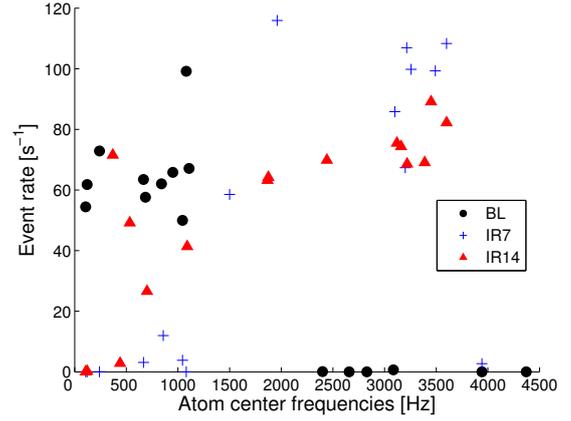}
\vspace*{-0.3cm}
\caption{
Scatter plot of atom event rates versus center frequencies of atoms for the BL, IR7 and IR14 cases.
The event rates are calculated during the last thirty minutes of vibration data in each case.
The introduction of a fault in the bearing leads to learning and activation of atoms with high center frequency.
}
\label{fig:EventHist}
\end{figure}

In \fig{EventHist} we present a scatter plot of atom event rates versus the center frequency for the three cases listed in \tab{centerfreq}.
It is evident that atoms with a lower center frequency occur in the BL case,
while the cases including a bearing fault (IR7 and IR14) result in adaptation and activation of atoms with higher center frequencies.
Furthermore, a comparison between the IR7 and IR14 cases reveals differences in the event rates associated with some of the atoms.
In summary, these results indicate that changes in the operational conditions and characteristics of a rotating machine
can be automatically detected using  unsupervised dictionary learning. 
Further work is required to investigate and develop reliable measures for change detection during continuous monitoring of a rotating machine,
including methods to avoid false positives associated with long-term variations in the operation of the machine.

\section{Discussion}

We investigate the possibility to automatically characterize a rotating machine
and detect when faults appear in the machine by monitoring a dictionary of learned atomic waveforms.
We find that the shape, frequency and repetition characteristics of the atoms depend on the operational conditions of the machine considered here.
Furthermore, we define the rate of change of atoms (the atom evolution rate) and illustrate that it can be useful for automatic detection of faults.
These results motivate further experiments with more realistic failure modes and varying operational conditions.
Further work is required to investigate and develop reliable measures for automatic change detection,
possibly using a complementary knowledge base including atoms learned from similar machines with known operational conditions.
In addition, deep learning extensions can be investigated for classification and prediction purposes.
Dictionary learning offers a novel approach to online condition monitoring,
which unlike most traditional techniques requires few assumptions about the machine and structure of the signal.
Further work in this direction is motivated in the search for condition monitoring methods that require little to none of configuration,
robust to changing operational conditions, and offers suitable scaling properties in the era of the Internet of Things.

\section{Acknowledgments}

This work is partially supported by SKF, the Kempe Foundations, and
the Swedish Foundation for International Cooperation in Research and Higher Education (STINT), grant number IG2011-2025.

\bibliographystyle{\refpath{IEEEbib}}
\bibliography{\refpath{dictev}}

\end{document}